\documentclass{article}

\usepackage{log_2026}						

\usepackage{booktabs}						
\usepackage{multirow}						
\usepackage{amsfonts}						
\usepackage{graphicx}						
\usepackage{duckuments}						

\usepackage[numbers,compress,sort]{natbib}	

\usepackage{graphicx}
\usepackage{booktabs}
\usepackage{subcaption}

\title[AttentionViG: Cross-Attention-Based Dynamic Neighbor Aggregation in Vision GNNs]{AttentionViG: Cross-Attention-Based Dynamic Neighbor Aggregation in Vision GNNs}

\author[Gedik et al.]{%
\textbf{Hakan Emre Gedik}$^{1}$ \quad
\textbf{Andrew Martin}$^{1}$ \quad
\textbf{Mustafa Munir}$^{1}$ \quad
\textbf{Oguzhan Baser}$^{1}$\\
\textbf{Radu Marculescu}$^{1}$ \quad
\textbf{Sandeep Chinchali}$^{1}$ \quad
\textbf{Alan Bovik}$^{1,2}$\\
$^{1}$The University of Texas at Austin \quad $^{2}$University of Colorado Boulder\\
\email{\{hakan.gedik,andrewm1177,mmunir,oguzhanbaser,radum,sandeepc\}@utexas.edu}\\
\email{Alan.Bovik@colorado.edu}
}

\begin{document}

\maketitle

\begin{abstract}
  Vision Graph Neural Networks (ViGs) have demonstrated promising performance in image recognition tasks against Convolutional Neural Networks (CNNs) and Vision Transformers (ViTs). An essential part of the ViG framework is the node-neighbor feature aggregation method. Although various graph convolution methods, such as Max-Relative, EdgeConv, GIN, and GraphSAGE, have been explored, a versatile aggregation method that effectively captures complex node-neighbor relationships without requiring architecture-specific refinements is needed. To address this gap, we propose a cross-attention-based aggregation method in which the query projections come from the node, while the key projections come from its neighbors. Additionally, we introduce a novel architecture called AttentionViG that uses the proposed cross-attention aggregation scheme to conduct non-local message passing. We evaluated the image recognition performance of AttentionViG on the ImageNet-1K benchmark, where it achieved SOTA performance. Additionally, we assessed its transferability to downstream tasks, including object detection and instance segmentation on MS COCO 2017, as well as semantic segmentation on ADE20K. Our results demonstrate that the proposed method not only achieves strong performance, but also maintains efficiency, delivering competitive accuracy with comparable FLOPs to prior vision GNN architectures.
\end{abstract}

\vspace{-3mm}

\section{Introduction}
With the advent of deep learning, unprecedented performance has been achieved in general computer vision tasks, including image classification \cite{imagenet}, object detection \cite{coco}, and image segmentation \cite{coco, ade20k}. Convolutional Neural Networks (CNNs), favored for their strong inductive bias and linear computational complexity with respect to input image resolution, formed the backbone of early deep learning models \cite{alexnet, resnet} and remain widely used \cite{convnet20s}. However, they are often criticized for their limited ability to capture global context, as convolutions are strictly local operators \cite{convnet20s}.

By contrast, Vision Transformers (ViTs) \cite{vit} offer global context modeling by partitioning each input image into non-overlapping patches and processing them as a sequence through Transformer layers \cite{attention}. Although ViTs outperform CNNs, they require large amounts of data and are more prone to overfitting due to their weaker inductive bias as compared to CNNs. Furthermore, the self-attention mechanism in Transformers has quadratic computational complexity with respect to the number of patches, which limits their scalability. Restricting the attention span to a predefined window \cite{swin} was introduced to enhance inductive bias and achieve linear computational scalability. However, this approach reduces the ability to model global context. In general, the design of ViTs involves a trade-off between computational complexity and global context modeling.

Vision GNNs (ViGs) introduced an unconventional approach to image representation by partitioning the input image into non-overlapping patches, representing each patch as a node, and forming a graph with all patches \cite{vig}. Since images do not inherently have a graph structure, a key aspect of ViG design is defining the policy that determines how nodes connect, and establishing neighbor relationships along with the function that aggregates information to nodes from their neighbors. In \cite{vig}, the nodes are connected to their k-nearest neighbors (kNN) based on their feature vectors. Although highly versatile, this policy can be criticized for the computational overhead of the kNN search, which scales quadratically with the number of patches in the input image. 

MobileViG \cite{mobilevig} was introduced to mitigate the computational cost of kNN search. Instead of dynamically determining each node's neighbors through an exhaustive search, it adopts a fixed graph construction with a criss-cross pattern. While MobileViG achieves strong performance and low latency in a low-parameter regime, its performance degrades as model size increases, as compared to other models with similar latency. This is mainly due to the fixed graph construction scheme, which forces nodes, regardless of their semantic relevance, to become neighbors. Supporting this, GreedyViG \cite{greedyvig} follows the same graph construction policy but improves performances across various parameter ranges by removing neighbors that do not meet a heuristic distance criterion. In general, graph construction policies that ignore the semantic relationships between nodes and their potential neighbors result in suboptimal performance \cite{mobilevig2}. 

A major reason for the sensitivity of performance to neighbor selection is the node-neighbor feature aggregation function. Typical graph convolution methods, such as Max-Relative \cite{mr}, EdgeConv \cite{edgeconv}, GIN \cite{gin}, and GraphSAGE \cite{graphsage}, lack a mechanism to assign importance weights to neighbors. For instance, the Max-Relative graph convolution, employed in vanilla ViG \cite{vig} and MobileViG, follows the formulation:
\begin{equation}
    \mathbf{x}_i' = \mathbf{W} \left[ \mathbf{x}_i, \max \left( \left\{ \mathbf{x}_j - \mathbf{x}_i \mid j \in \mathcal{N}(\mathbf{x}_i) \right\} \right) \right].
    \label{eq:mr}
\end{equation}
\noindent  where $x_i$ is the feature vector of the $i$'th node, $\mathcal{N}(\mathbf{x}_i)$ denotes the set of feature vectors of its neighbors, [., .] indicates feature-wise concatenation, $\mathbf{W}$ is a learnable linear projection, and $x_i'$ is the aggregated feature. We argue that neighbors semantically unrelated to $x_i$ in $\mathcal{N}(x_i)$ act as noise through the max operation. Although the dynamic graph construction in vanilla ViG compensates for the simplicity of the aggregation function, when the graph construction policy is fixed or imperfect in any way, it negatively impacts model performance.

To address this issue, we propose a general-purpose aggregation method based on cross-attention, where the query is derived from a node and the keys from its neighbors. The query-key cosine similarities are converted into attention scores over the neighbors using an exponential kernel, without enforcing competition among them. The neighbor features are first projected using value functions, then weighted by the attention scores, concatenated with the node's own features, and passed through a learnable linear transformation followed by a nonlinearity to produce the final output.

Furthermore, we introduce AttentionViG, a ViG architecture comprising of inverted residual blocks \cite{mobilenetv2} and Grapher layers that implement our proposed cross-attention aggregation function. To conduct graph construction, we adopt Sparse Vision Graph Attention (SVGA) \cite{mobilevig} due to its low computational cost, even though its fixed connectivity assigns neighbors to nodes without considering semantic relationships. We show that cross-attention aggregation mitigates the limitations of SVGA, enabling AttentionViG to achieve SOTA performance on ImageNet-1k \cite{imagenet} classification, object detection and instance segmentation on MS-COCO \cite{coco}, and semantic segmentation on ADE20K \cite{ade20k}. Our main contributions are as follows: 
\begin{enumerate}
    \item We propose a general-purpose cross-attention-based feature aggregation method for graph neural networks (GNNs). Our aggregation function learns to weigh the contribution of each neighbor, effectively discarding irrelevant ones and enhancing message passing. 
    \item We design a multi-scale ViG architecture, AttentionViG, which consists of inverted residual blocks for local processing and Grapher layers that apply the proposed cross-attention aggregation on SVGA graph construction for non-local message passing.      
    \item Through extensive experiments, we show that our model outperforms existing CNNs, ViTs, and ViGs across various model sizes on ImageNet-1k classification, object detection and instance segmentation on COCO, and semantic segmentation on ADE20K.  
\end{enumerate} 

The rest of this paper is organized as follows. In Section \ref{sec:related_work}, we summarize prominent CNNs, ViTs, and ViGs in image recognition. Section \ref{sec:methodology} details the cross-attention aggregation and the AttentionViG model. Section \ref{sec:experiments} highlights the experimental setup and establishes the performance of AttentionViG on ImageNet-1k image classification, MS-COCO object detection and instance segmentation, and ADE20K semantic segmentation, in comparison aginst other SOTA models. Lastly, we summarize our contributions and discuss future work in Section \ref{sec:conclusion}. 
\vspace{-3mm}
\section{Related Work}
\vspace{-1mm}

\label{sec:related_work} 
\begin{figure}[t]
\centering
\begin{subfigure}[t]{0.40\linewidth}
    \centering
    \includegraphics[width=\linewidth]{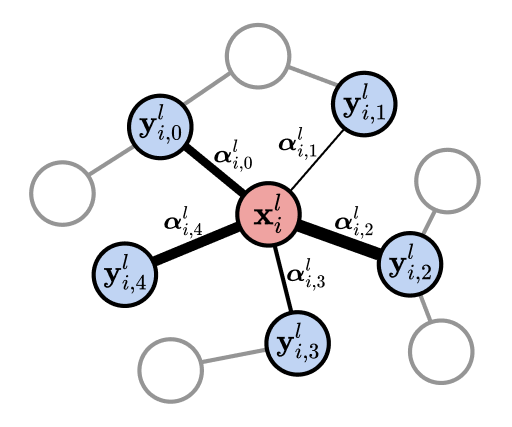}
    \caption{Cross-attention assigns weights to neighboring nodes, with thicker edges indicating larger weights. For node $\mathbf{x}_i^l$ and its neighbors $\mathbf{y}_{i,0}^l, \mathbf{y}_{i,1}^l, \mathbf{y}_{i,2}^l, \mathbf{y}_{i,3}^l$, the corresponding weights are $\alpha_{i,0}^l, \alpha_{i,1}^l, \alpha_{i,2}^l, \alpha_{i,3}^l$.}
    \label{fig:neighbor_weights}
\end{subfigure}
\hfill
\begin{subfigure}[t]{0.40\linewidth}
    \centering
    \includegraphics[width=\linewidth]{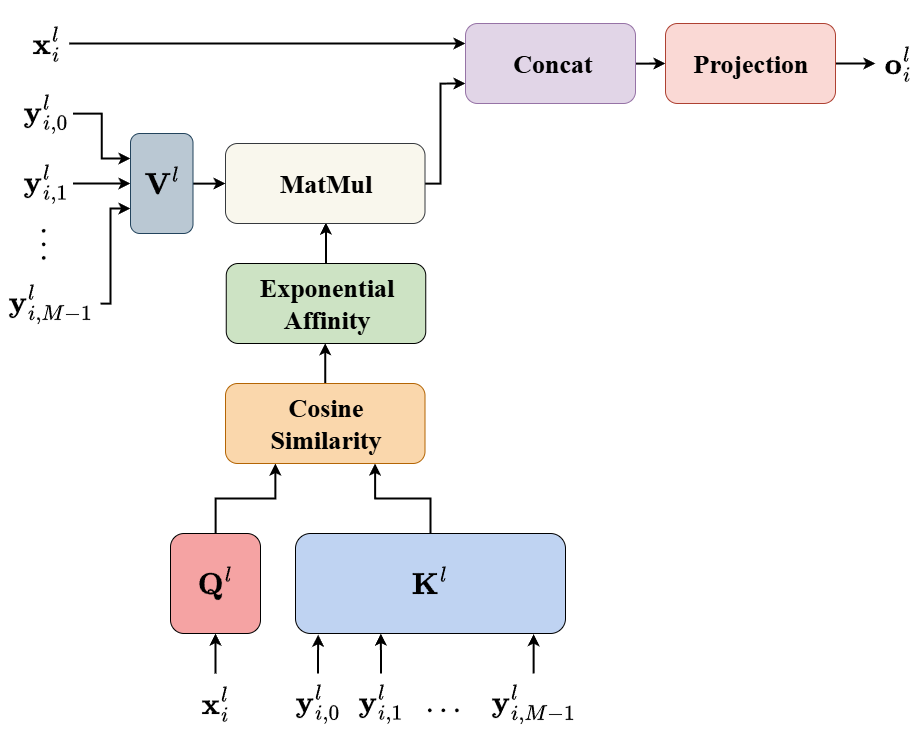}
    \caption{Cross-attention-based feature aggregation extracts query vectors from nodes and key vectors from neighboring nodes, enabling the model to learn the relative importance of each neighbor.}
    \label{fig:cross_attention}
\end{subfigure}

\caption{Illustration of cross-attention-based neighbor weighting and feature aggregation in the proposed graph framework.}
\label{fig:cross_attention_framework}
\vspace{-5mm}
\end{figure} 
The paradigm in image recognition model design underwent a significant shift with the introduction of AlexNet \cite{alexnet}, which built upon the foundation established by LeNet-1 \cite{lenet}, the first convolutional model. Since then, a variety of CNN architectures have been proposed, with a focus on performance \cite{vgg, resnet, regnet, densenet, convnet20s} or efficiency \cite{squeezenet, mobilenet, mobilenetv2, shufflenet, efficientnet, efficientnetv2, rapidnet}. Due to their inherent inductive biases, such as shift-invariance and locality, CNNs remain widely used in computer vision.

Originally developed for sequence modeling, Transformers \cite{attention} were first introduced to computer vision in \cite{vit}, demonstrating their potential as an alternative to convolutional networks. While they have a weaker inductive bias than CNNs, their global receptive field enables them to model long-range dependencies more effectively, often leading to superior performance \cite{deit, crossvit, t2t, pvt, mvt}. However, Transformers are computationally expensive and require large amounts of data. To address these challenges, several approaches have reintegrated convolutional layers, leading to more efficient lightweight models with improved performance \cite{cvt, mobilevit, efficientformer, efficientformerv2}. 

Graph Neural Networks (GNNs) were introduced as a general-purpose backbone for vision tasks in \cite{vig}, employing a kNN search for dynamic graph construction, which introduced a significant computational bottleneck. MobileViG \cite{mobilevig} mitigated this issue by proposing SVGA, a static graph construction method that reduces computational cost. GreedyViG \cite{greedyvig} further improved upon SVGA with Dynamic Axial Graph Construction (DAGC), refining neighbor selection by discarding SVGA-assigned neighbors that do not meet a heuristic distance criterion.

Other graph construction policies have also been explored. LogViG \cite{logvig} introduced an exponentially increasing neighbor distance to replace the fixed strides in SVGA, limiting the number of global connections, especially for large images. SViG replaced kNN selection in \cite{vig} with similarity-based thresholding, enabling nodes to have a variable number of neighbors. WiGNet \cite{wignet} proposed partitioning the input image into non-overlapping windows and constructing a graph within each window, achieving linear computational complexity with respect to input resolution, in contrast to the quadratic complexity in \cite{vig}. Similarly, ClusterViG \cite{clustervig} introduced a partitioning scheme that confines kNN searches within partitions. This restriction improves parallelism and significantly accelerates inference compared to \cite{vig}. ViHGNN \cite{hgnn} utilized hypergraphs, a generalization of graphs, to create a more expressive model capable of capturing intricate relationships among nodes.  

At the aggregation level, recent vision GNNs have also explored adaptive weighting at different structural levels. HgVT \cite{hgvt} employs softmax-based fuzzy cross-attention between vertices and hyperedges, ClusterViG \cite{clustervig} uses GATv2 to aggregate global partition-level features before local GCN updates, and DVHGNN \cite{dvhgnn} adaptively weights hypergraph messages using sigmoid-gated cosine similarities and learned dilation-specific weights. While these approaches incorporate adaptive aggregation within specialized partitioned or hypergraph designs, our method directly aggregates node–neighbor features and can be applied independently of the underlying graph construction policy. We demonstrate that even with the relatively simple SVGA graph construction policy, AttentionViG achieves SOTA performance in image classification, object detection, and segmentation tasks.

\vspace{-3mm}
\section{Methodology}
\label{sec:methodology} 
\begin{figure*}[t!]
    \centering
    \includegraphics[width=0.80\textwidth]{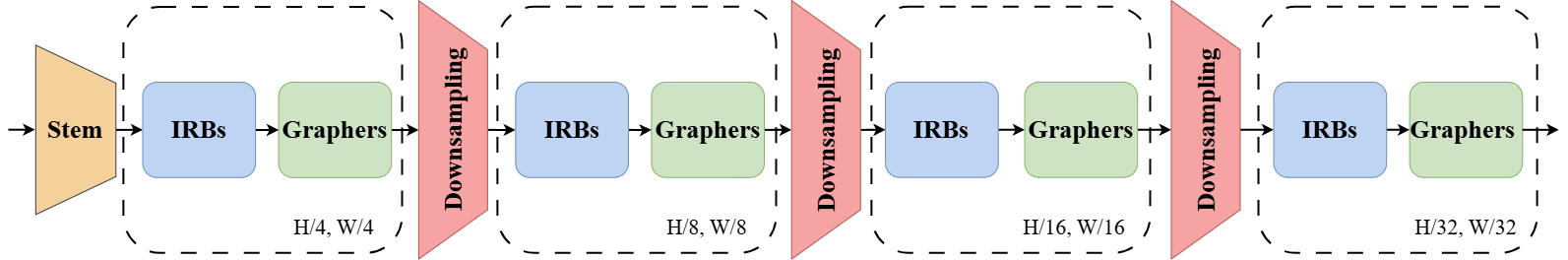}
    \caption{The overall architecture of AttentionViG consists of a stem, inverted residual blocks (IRB) for feature extraction, Grapher layers for graph-based feature aggregation, and downsampling blocks for multi-scale representation learning.}
    \label{fig:architecture}
\end{figure*} 
\begin{figure}[t!]
    \centering
    \includegraphics[width=0.50\textwidth]{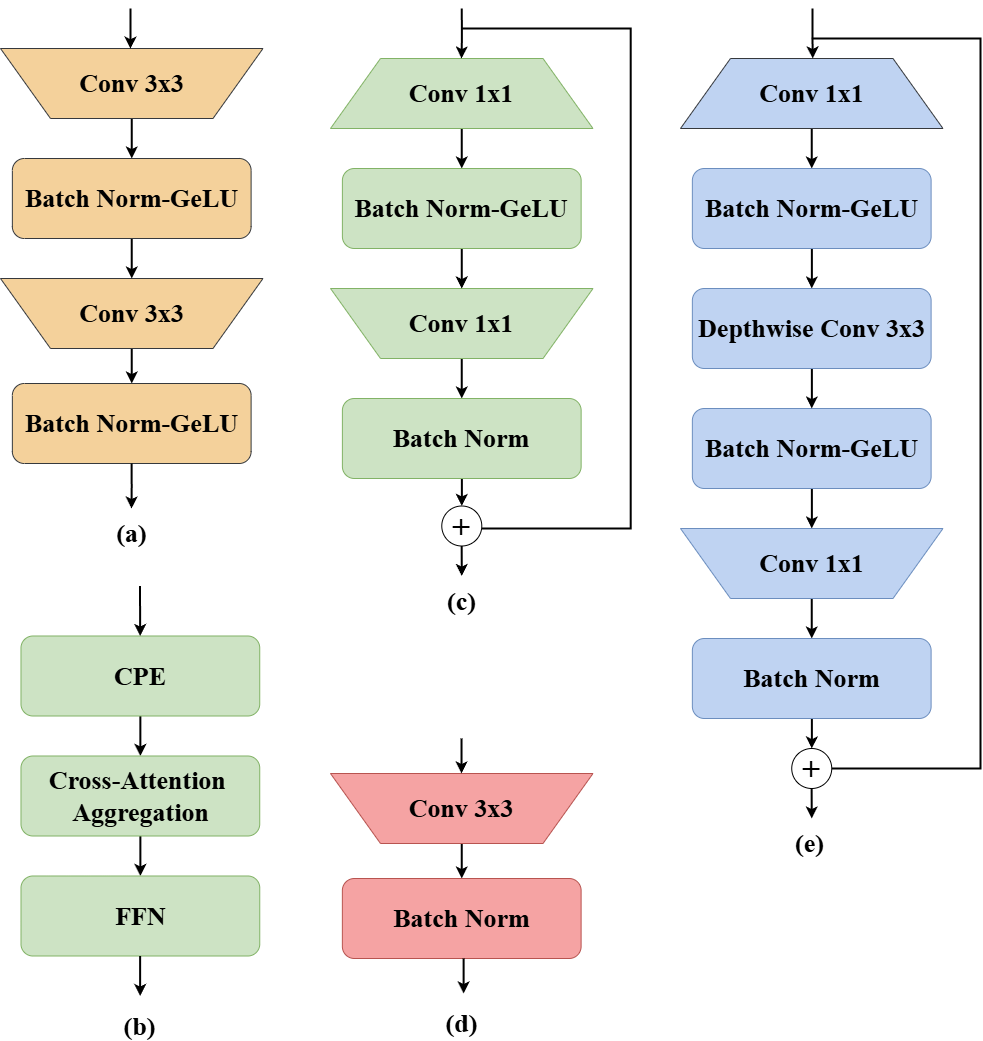}
    \caption{(a) Convolutional stem for input image embeddings, where convolutional layers have a stride of 2. (b) Grapher layer with CPE \cite{conditionalpe} and the proposed cross-attention aggregation. (c) FFN layer, a component of the Grapher. (d) Downsampling block with a convolutional layer of stride 2. (e) Inverted residual block as introduced in \cite{mobilenetv2}.}
    \label{fig:blocks}
    \vspace{-3mm}
\end{figure} 

\subsection{Cross-Attention Aggregation} 
\label{sec:cross-attention} 
Widely used aggregation functions, such as Max-Relative, lack a mechanism to assign weights to neighbors, treating them all equally. Consequently, replacing the kNN in vanilla ViG with a more efficient yet imprecise graph construction method, as discussed in Section \ref{sec:related_work}, may lead to suboptimal performance. To address this issue, we propose a cross-attention-based aggregation function that dynamically determines the relevance of each neighbor to the node. Specifically, we first divide the input image into non-overlapping patches and process them with a convolutional stem to obtain the initial node representations for $l=0$, denoted as $\{\mathbf{x_0}^l, \dots, \mathbf{x}_{N-1}^l\}$, where $l$ represents the GNN layer, and $\mathbf{x}_i^l$ is the feature vector of the $i^{th}$ node in the $l^{th}$ GNN layer for $i \in \{0, \dots, N-1\}$.       

For each node $\mathbf{x}_i^l$, the neighbor set $\mathcal{N}(\mathbf{x}_i^l)$ is sampled based on a predefined graph construction policy. Then, query vectors are derived from the node vectors, while key vectors are obtained from their corresponding neighbor vectors as follows: 
\begin{equation}
    \mathbf{q}_i^l = \mathbf{Q}^l \mathbf{x}_i^l      
    \label{eq:query}
\end{equation}
\begin{equation}
    \mathbf{k}_{i, j}^l = \mathbf{K}^l \mathbf{y}_{i, j}^l   
    \label{eq:key}
\end{equation}
\noindent where $\mathbf{y}_{i, j}^l \in \mathcal{N}(\mathbf{x}_i^l)$ for $j \in \{0 \dots M-1\}, M=|\mathcal{N}(\mathbf{x}_i^l)|$, and $\mathbf{Q}^l$, $\mathbf{K}^l$ are learnable linear projections. The relevance score $\mathbf{s}_{i, j}^l$ between a node and its neighbor is computed as the cosine similarity of their respective query and key vectors:
\begin{equation}
    \mathbf{s}_{i, j}^l = \frac{(\mathbf{q}_i^l)^T \mathbf{k}_{i, j}^l}{\|\mathbf{q}_i^l\|_2 \cdot \|\mathbf{k}_{i, j}^l\|_2}
    \label{eq:score}
\end{equation}
\noindent To transform this similarity into an attention weight, an exponential kernel with a learnable scaling parameter $\beta$ is applied:  
\begin{equation}
    \alpha_{i, j}^l = \exp\left(-\beta \left(1 - \mathbf{s}_{i, j}^l\right)\right)
    \label{eq:exp_affinity}
\end{equation}
For aggregation, the neighbors' feature vectors are first projected through a learnable linear projection:   
\begin{equation}
    \mathbf{v}_{i, j}^l = \mathbf{V}^l \mathbf{y}_{i, j}^l      
    \label{eq:value}
\end{equation}
Finally, the output is computed by concatenating the node feature vector with the attention-weighted sum of its corresponding value vectors, followed by a non-linearity and a linear projection:   
\begin{equation}
    \mathbf{o}_i^l = \sigma(\mathbf{W} [\mathbf{x}_i^l, \sum_j \mathbf{\alpha}_{i, j}^l \mathbf{v}_{i, j}^l] )
    \label{eq:output}
\end{equation}
\noindent where [·, ·] denotes feature-wise concatenation, $\mathbf{W}$ is a learnable projection, and $\sigma$ is a nonlinearity, for which we use GeLU. 

The exponential kernel in equation \ref{eq:exp_affinity}, referred to as the exponential affinity function, was introduced in Tip-Adapter \cite{tip-adapter} as a similarity measure between learned keys and queries. Unlike softmax, it does not enforce competition among neighbors, allowing for more flexible attention aggregation. The exponential affinity function empirically outperforms softmax, as demonstrated later in Section \ref{sec:ablation}.

Cross-attention aggregation shares some similarities with Graph Attention Networks (GAT) \cite{gat} in that both approaches compute attention scores between nodes and their neighbors. However, our approach differs significantly from GAT. In particular, GAT can be formulated as follows:
\begin{equation}
    \mathbf{s}_{i, j}^l = \sigma(\mathbf{a}^T[\mathbf{W}\mathbf{x}_i^l, \mathbf{W}\mathbf{y}_{i, j}^l]) 
    \label{eq:score_gat}
\end{equation}
\begin{equation}
    \alpha_{i, j}^l = \frac{\exp(\mathbf{s}_{i, j}^l)}{\sum_k \exp(\mathbf{s}_{i, k}^l)}, 
    \label{eq:softmax}
\end{equation}
\begin{equation}
    \mathbf{o}_i^l = \sigma(\sum_j \mathbf{\alpha}_{i, j}^l \mathbf{W}\mathbf{y}_{i, j}^l)
    \label{eq:output_gat}
\end{equation}
\noindent where $\sigma$ is a nonlinearity, and $\mathbf{W}$ and $\mathbf{a}$ are learnable linear projections. GAT utilizes softmax, as in Eq. \ref{eq:softmax}, to obtain $\mathbf{\alpha}^l_{i, j}$. Comparing Eqs. \ref{eq:score} and \ref{eq:score_gat}, GAT uses a shared node-neighbor projection along with a nonlinearity, whereas cross-attention aggregation learns separate query and key projections for nodes and neighbors. Additionally, comparing Eqs. \ref{eq:output} and \ref{eq:output_gat}, GAT outputs weighted and projected neighbor features, while cross-attention aggregation learns a projection for the weighted and aggregated neighbors along with the node.

In summary, the cross-attention aggregation functions as a dynamic feature mixer for the neighbors, with learnable weights that determine each neighbor's relevance to its node. Our method differs significantly from previously proposed aggregation methods such as Max-Relative, EdgeConv, GIN, GraphSAGE, and GAT.

\vspace{-1mm}
\subsection{Grapher Layer}
\vspace{-1mm}
Using the proposed cross-attention aggregation, we design our Grapher layer. Specifically, the Grapher layer consists of the cross-attention aggregation, as described in Eq. \ref{eq:output}, followed by a feed-forward network (FFN) with a single hidden layer, a residual connection, GeLU activation \cite{gelu}, and an expansion ratio of 4. Additionally, the Grapher layer incorporates conditional positional encoding (CPE) \cite{conditionalpe}, as applied in \cite{greedyvig}. Overall, the Grapher layer can be expressed as:
\begin{equation}
    \mathbf{g}_i^l = \mathbf{FFN}^l(\mathbf{Aggregation}^l(\mathbf{CPE}^l(\mathbf{x}_i^l)))   
    \label{eq:grapher_output}
\end{equation}
We adopt CPE to encode spatial positional information for the neighbors, as other fixed or learnable positional encodings \cite{attention} may result in poor generalization across spatial resolutions not encountered during training. Combined with a global graph construction policy, the cross-attention aggregation, and the FFN layer, the grapher functions as a nonlinear, global feature mixer. 

\vspace{-3mm}
\subsection{AttentionViG Architecture} 
\label{sec:architecture} 
\vspace{-1mm}
As illustrated in Figure \ref{fig:architecture}, AttentionViG is a multiscale hybrid GNN-CNN network comprising a convolutional stem for image embeddings, inverted residual blocks \cite{mobilenetv2} for local processing, and Grapher layers using the SVGA \cite{mobilevig} graph construction policy for global message passing. 

SVGA connects nodes that are horizontally or vertically aligned with a stride of 2. The SVGA policy maintains a fixed connectivity pattern across layers. Compared to kNN search \cite{vig}, it is computationally more efficient. However, its static nature prevents it from filtering out semantically irrelevant nodes, a limitation addressed by the neighbor weighting mechanism in the proposed cross-attention aggregation.

The convolutional stem consists of two convolutional layers with a kernel size of $3\times3$ and a stride of 2, each followed by batch normalization \cite{batchnorm} and GeLU activation. For inverted residual blocks, we use GeLU activation and set the expansion ratio to 4. AttentionViG operates on four scales, with downsampling performed using a convolutional layer with a kernel size of $3\times3$ and a stride of 2, followed by batch normalization. The number of inverted residual blocks in each stage varies based on the model size, while the number of Grapher layers is fixed at two. This choice is based on the properties of SVGA, where cascading two Grapher layers is sufficient to achieve global message passing, while a single layer is inadequate and additional layers may be redundant. More details on the network configurations are provided in Supplementary Materials. 

Overall, AttentionViG is a hybrid CNN-GNN architecture that uses inverted residual blocks for local processing and Grapher layers with the SVGA graph construction policy for global processing. With its parallelized Grapher implementation, the model's computational complexity scales linearly with input image resolution, ensuring efficiency on the image recognition task.  
\vspace{-3mm}
\section{Experimental Results} 
\label{sec:experiments} 
\begin{table}[!t]
    \centering
    \renewcommand{\arraystretch}{0.80}
    \caption{Compared SOTA models on ImageNet-1k classification. Accuracy results may vary by approximately 0.2\% across different training seeds. The number of parameters is given in millions (M). A dash (-) indicates missing data in the original papers.} 
    \begin{tabular}{l c c c c}
        \toprule
        \textbf{Model} & \textbf{Type} & \textbf{Params. (M)} & \textbf{GFLOPs} & \textbf{Accuracy (\%)} \\
        \midrule
        ResNet50 \cite{resnet} & CNN & 25.6 & 4.1  & 80.4 \\
        RegNetY-16GF \cite{regnet} & CNN & 83.6 & 15.9  & 80.4 \\
        ConvNext-T \cite{convnet20s} & CNN & 28.6 & 7.4 & 82.7 \\
        ConvNext-S \cite{convnet20s} & CNN & 50.0 & 12.9  & 83.1 \\
        \midrule
        EfficientFormer-L1 \cite{efficientformer} & CNN-ViT & 12.3 & 1.3  & 79.2 \\
       EfficientFormerV2-L \cite{efficientformerv2} & CNN-ViT & 26.1 & 2.6  & 83.3 \\
        LeViT-192 \cite{levit} & CNN-ViT & 10.9 & 0.7  & 80.0 \\
        LeViT-384 \cite{levit} & CNN-ViT & 39.1 & 2.4  & 82.6 \\
        \midrule
        PVT-Small \cite{pvt} & ViT & 24.5 & 3.8  & 79.8 \\
        PVT-Large \cite{pvt} & ViT & 61.4 & 9.8  & 81.7 \\
        Swin-T \cite{swin} & ViT & 29.0 & 4.5  & 81.3 \\
        Swin-S \cite{pvt} & ViT & 50.0 & 8.7  & 83.0 \\
        CrossViT-15 \cite{crossvit} & ViT & 27.4 & 5.8  & 81.5 \\
        CrossViT-18 \cite{crossvit} & ViT & 43.3 & 9.0  & 82.5 \\
        \midrule
        Vim-S \cite{vim} & Mamba & 26.0 & -  & 80.5 \\
        EfficientVMamba-S \cite{efficient-mamba} & Mamba & 11.0 & 1.3  & 78.7 \\
        EfficientVMamba-B \cite{efficient-mamba} & Mamba & 33.0 & 4.0  & 81.8 \\
        MambaVision-T \cite{mamba-vision} & Mamba & 31.8 & 4.4  & 82.3 \\
        MambaVision-S \cite{mamba-vision} & Mamba & 50.1 & 7.5  & 83.3 \\
        VMamba-T \cite{vmamba} & Mamba & 30.0 & 4.9  & 82.6 \\
        VMamba-S \cite{vmamba} & Mamba & 50.0 & 8.7  & 83.6 \\
        \midrule
        PViG-Ti \cite{vig} & GNN & 10.7 & 4.3  & 78.2 \\ 
        PViG-S \cite{vig} & GNN & 27.3 & 4.6  & 82.1 \\ 
        PViG-M \cite{vig} & GNN & 51.7 & 9.0  & 83.1 \\ 
        \midrule
        MobileViG-M \cite{mobilevig} & CNN-GNN & 14.0 & 1.5  & 80.6 \\ 
        MobileViG-B \cite{mobilevig} & CNN-GNN & 26.7 & 2.8  & 82.6 \\
        HgVT-S \cite{hgvt} & CNN-GNN & 22.9 & 5.5  & 81.2 \\ 
        DVHGNN-T \cite{dvhgnn} & CNN-GNN & 11.1 & 1.9  & 79.8 \\ 
        DVHGNN-S \cite{dvhgnn} & CNN-GNN & 30.2 & 5.2  & 83.1 \\ 
        PVG-S \cite{pvg} & CNN-GNN & 22.0 & 5.0  & 83.0 \\ 
        PVG-M \cite{pvg} & CNN-GNN & 42.0 & 8.9  & 83.7 \\ 
        CViG-Ti \cite{clustervig} & CNN-GNN & 11.5 & 1.3  & 80.3 \\ 
        CViG-S \cite{clustervig} & CNN-GNN & 28.2 & 4.2  & 83.7 \\
        WiGNet-Ti \cite{wignet} & CNN-GNN & 10.8 & 2.1 & 78.8 \\ 
        WiGNet-S \cite{wignet} & CNN-GNN & 27.4 & 5.7 & 82.0 \\ 
      GreedyViG-S \cite{greedyvig} & CNN-GNN & 12.0 & 1.6  & 81.1 \\ 
        GreedyViG-M \cite{greedyvig} & CNN-GNN & 21.9 & 3.2  & 82.9 \\ 
      GreedyViG-B \cite{greedyvig} & CNN-GNN & 30.9 & 5.2  & 83.9 \\ 
        \midrule
        \textbf{AttentionViG-S (Ours)} & \textbf{CNN-GNN} & \textbf{12.3} & \textbf{1.6} & \textbf{81.3} \\
        \textbf{AttentionViG-M (Ours)} & \textbf{CNN-GNN} & \textbf{22.2} & \textbf{3.2} & \textbf{83.1} \\
        \textbf{AttentionViG-B (Ours)} & \textbf{CNN-GNN} & \textbf{32.3} & \textbf{4.8} & \textbf{83.9} \\
        \bottomrule
    \end{tabular}
    \label{tab:classification}
\end{table}

\begin{table}[!t]
    \centering
    \footnotesize
    \renewcommand{\arraystretch}{0.80}
    \caption{Compared SOTA models on MS-COCO 2017 object detection/instance segmentation and ADE20k semantic segmentation. The number of parameters is given in millions (M). A dash (-) indicates missing data in the original papers.}
    \setlength{\tabcolsep}{2pt}
    \begin{tabular}{lcccccccc}
        \toprule
        \textbf{Backbone} & \textbf{Params.} & $AP^{box}$ & $AP^{box}_{50}$ & $AP^{box}_{75}$ & $AP^{mask}$ & $AP^{mask}_{50}$ & $AP^{mask}_{75}$ & $mIoU$ \\
        \midrule
        ResNet18 \cite{resnet} & 11.7 & 34.0 & 54.0 & 36.7 & 31.2 & 51.0 & 32.7 & 32.9 \\
        EfficientFormer-L1 \cite{efficientformer} & 12.3 & 37.9 & 60.3 & 41.0 & 35.4 & 57.3 & 37.3 & 38.9 \\
        EfficientFormerV2-S2 \cite{efficientformerv2} & 12.6 & 43.4 & 65.4 & 47.5 & 39.5 & 62.4 & 42.2 & 42.4 \\
        PoolFormer-S12 \cite{poolformer} & 12.0 & 37.3 & 59.0 & 40.1 & 34.6 & 55.8 & 36.9 & 37.2 \\
        FastViT-SA12 \cite{fastvit} & 10.9 & 38.9 & 60.5 & 42.2 & 35.9 & 57.6 & 38.1 & 38.0 \\
        MobileViG-M \cite{mobilevig} & 14.0 & 41.3 & 62.8 & 45.1 & 38.1 & 60.1 & 40.8 & - \\
        GreedyViG-S \cite{greedyvig} & 12.0 & 43.2 & 65.2 & 47.3 & 39.8 & 62.2 & 43.2 & 43.2 \\
        
        \textbf{AttentionViG-S} & 12.3 & \textbf{43.5} & \textbf{65.8} & \textbf{47.6} & \textbf{40.0} & \textbf{62.8} & \textbf{43.1} & \textbf{43.8} \\
        \midrule
        EfficientFormer-L7 \cite{efficientformer} & 82.1 & 42.6 & 65.1 & 46.1 & 39.0 & 62.2 & 41.7 & 45.1 \\
        EfficientFormerV2-L \cite{efficientformerv2} & 26.1 & 44.7 & 66.3 & 48.8 & 40.4 & 63.5 & 43.2 & 45.2 \\
        PoolFormer-S24 \cite{poolformer} & 21.0 & 40.1 & 62.2 & 43.4 & 37.0 & 59.1 & 39.6 & 40.3 \\
        FastViT-SA36 \cite{fastvit} & 30.4 & 43.8 & 65.1 & 47.9 & 39.4 & 62.0 & 42.3 & 42.9 \\
        Swin-T \cite{swin} & 29.0 & 43.7 & 66.6 & 47.7 & 39.8 & 63.3 & 42.7 & 43.1 \\
        PViG-S \cite{vig} & 27.3 & 42.6 & 65.0 & 46.0 & 39.4 & 62.4 & 41.6 & - \\
        PViHGNN-S \cite{vihgnn} & 28.5 & 43.1 & 66.0 & 46.5 & 39.6 & 63.0 & 42.3 & - \\
        PVT-Small \cite{pvt} & 24.5 & 40.4 & 62.9 & 43.8 & 37.8 & 60.1 & 40.3 & 39.8 \\
        MobileViG-B \cite{mobilevig} & 26.7 & 42.0 & 64.3 & 46.0 & 38.9 & 61.4 & 41.6 & - \\
        PVG-S \cite{pvg} & 22.0 & 43.9 & 66.3 & 48.0 & 39.8 & 62.8 & 42.4 & - \\
        DVHGNN-S \cite{dvhgnn} & 30.2 & 44.8 & 66.8 & 49.0 & 40.2 & 63.5 & 43.1 & 46.8 \\
        GreedyViG-B \cite{greedyvig} & 30.9 & 46.3 & 68.4 & 51.3 & 42.1 & 65.5 & 45.4 & 47.4 \\
        CViG-S \cite{clustervig} & 28.2 & 47.4 & 68.1 & 52.0 & 43.4 & 67.2 & 47.5 & - \\
        VMamba-T \cite{vmamba} & 30.0 & 47.3 & - & - & 42.7 & - & - & \textbf{47.9} \\
            MambaVision-T \cite{mamba-vision} & 31.8 & \textbf{51.1} & \textbf{70.0} & \textbf{55.6} & \textbf{44.3} & \textbf{67.3} & \textbf{47.9} & 46.0 \\
        \textbf{AttentionViG-B} & 32.3 & 46.4 & 68.5 & 51.3 & 42.3 & 65.5 & 45.6 & 47.8 \\
        \bottomrule
    \end{tabular}
    \label{tab:detection}
\end{table} 

\vspace{-1mm}
\subsection{Imagenet-1k Classification}
\label{sec:classification} 
\vspace{-1mm}

We trained our model and report top-1 validation accuracy on the ImageNet-1K \cite{imagenet} dataset. The experiments were conducted using 16 NVIDIA A100 GPUs with a batch size of 2048 over 300 epochs. We employed the AdamW optimizer \cite{adamw} with cosine annealing, starting with an initial learning rate of $2 \times 10^{-3}$. The input image resolution was fixed at $224\times224$. When training the classification model, we applied hard knowledge distillation \cite{deit} using RegNetY-16GF \cite{regnet} as the teacher model. To conduct data augmentation, we followed the method used in \cite{vig, vihgnn}.  

As shown in Table \ref{tab:classification}, AttentionViG models outperform PViG \cite{vig}, MobileViG \cite{mobilevig}, CViG \cite{clustervig}, WiGNet \cite{wignet}, EfficientFormer \cite{efficientformer}, GreedyViG \cite{greedyvig}, and other comparable convolutional or hybrid models in terms of parameters and FLOPs. Our smallest model achieves 81.3\% top-1 accuracy, surpassing PViG-Ti by 3.1\%, and GreedyViG by 0.2\%, while using equal or fewer FLOPs. 

Scaling up further improves performance, with our largest model reaching 83.9\% top-1 accuracy, surpassing PViG, DVHGNN \cite{dvhgnn}, PVG \cite{pvg},  CrossViT \cite{crossvit}, Swin \cite{swin}, and the EfficientFormer family \cite{efficientformer, efficientformerv2}, even when some of these models have nearly twice as many parameters. These results highlight the effectiveness of AttentionViG’s graph-based feature aggregation in enhancing classification accuracy while maintaining efficiency.

\subsection{Object Detection and Instance Segmentation}
\label{sec:coco} 

We evaluated AttentionViG’s generalizability on object detection and instance segmentation using the MS-COCO 2017 dataset \cite{coco}. We adopted the Mask R-CNN \cite{maskrcnn} framework with AttentionViG, pretrained on ImageNet-1K, as the backbone. The model was trained over 12 epochs with a batch size of 16 using the AdamW \cite{adamw} optimizer, an initial learning rate of $2 \times 10^{-2}$, and a weight decay of $0.05$. The learning rate was reduced by a factor of 10 at epochs 8 and 11. The input image resolution was fixed at $1333 \times 800$. 

As shown in Table \ref{tab:detection}, our smallest model achieved $AP^{box} = 43.5$ and $AP^{mask} = 40.0$ for object detection and instance segmentation, respectively, outperforming EfficientFormer \cite{efficientformer}, PoolFormer \cite{poolformer}, FastViT \cite{fastvit}, MobileViG \cite{mobilevig}, and GreedyViG \cite{greedyvig}. This demonstrates that AttentionViG maintains strong performance even in resource-constrained settings.

Scaling up the model further improved performance. Our larger model, AttentionViG-B, achieved $AP^{box} = 46.4$ and $AP^{mask} = 42.3$, surpassing EfficientFormerV2-L \cite{efficientformerv2}, PViG \cite{vig}, PVT-Small \cite{pvt}, and GreedyViG \cite{greedyvig}. Notably, despite having a similar parameter count to PViHGNN-S, AttentionViG-B consistently achieved better results across all metrics, demonstrating the effectiveness of its graph-based feature aggregation in enhancing object detection and instance segmentation.

\vspace{-1mm}
\subsection{Semantic Segmentation}
\label{sec:ade20k}
\vspace{-1mm}

We also evaluated the generalizability of AttentionViG on the semantic segmentation task using the ADE20K dataset \cite{ade20k}. We adopted the panoptic segmentation framework semantic FPN \cite{semanticfpn} with AttentionViG, pretrained on ImageNet-1K, as the backbone. The model was trained for $40,000$ iterations with a batch size of 32 using the AdamW \cite{adamw} optimizer, an initial learning rate of $2 \times 10^{-4}$, a weight decay of $10^{-4}$, and poly learning rate decay with a power of $0.9$. The input image resolution was fixed at $512 \times 512$. As in the object detection and instance segmentation experiments, the $\beta$ values in Eq. \ref{eq:exp_affinity} were frozen. 
 
As shown in Table \ref{tab:detection}, the smallest AttentionViG achieved an mIoU of 43.8, outperforming EfficientFormer \cite{efficientformer}, EfficientFormerV2 \cite{efficientformerv2}, FastViT \cite{fastvit}, and PoolFormer \cite{poolformer}. Our largest model further improved performance, achieving an mIoU of 47.8, demonstrating that AttentionViG scales effectively for dense prediction tasks while maintaining competitive efficiency in terms of parameters and FLOPs.

These results show that AttentionViG performs competitively in semantic segmentation, with consistent gains across scales, indicating the effectiveness of its feature aggregation for both classification and dense prediction.

\vspace{-1mm}
\subsection{Visualization of Neighbor Weights} 
\vspace{-1mm}

\begin{figure}[t!]
    \centering
    \includegraphics[width=0.40\textwidth]{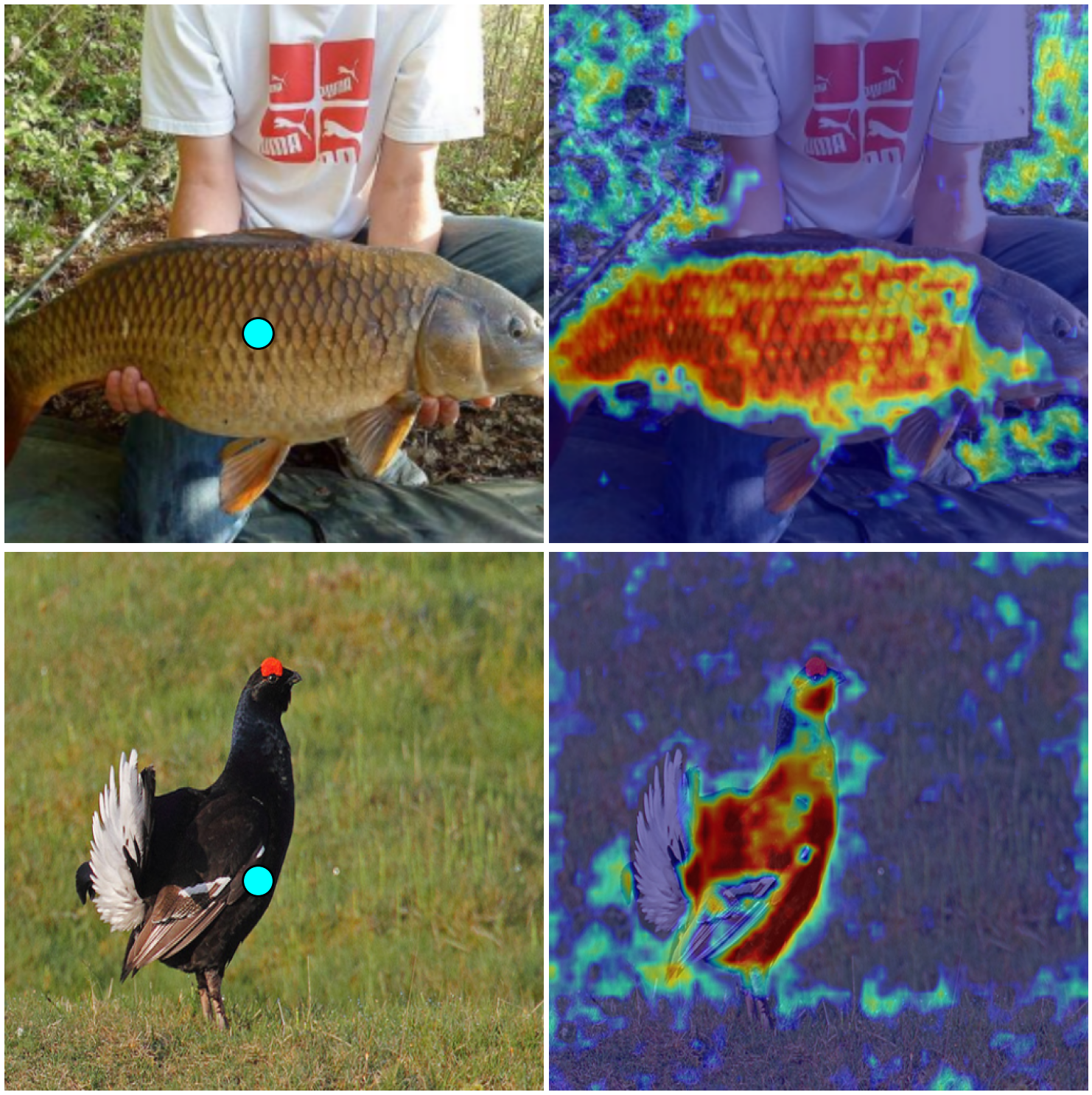}
    \caption{Query-key cosine similarity visualization. The query is from the cyan point (left); the heatmap overlays cosine similarity across the image, with warmer regions indicating higher similarity.} 
    \label{fig:heatmap}
    \vspace{-3mm}
\end{figure}

In Figure \ref{fig:heatmap}, we visualized the learned neighbor weights in the second Grapher layer. For selected locations highlighted in cyan, we computed the cosine similarity between the learned query vector at that location and the learned key vectors across the image, and overlaid the resulting heatmap on the original image.

We observed that the model amplifies neighbors that are semantically related to the query location while mostly suppressing unrelated regions. This suggests that attention can compensate for imperfections in graph construction by dynamically assigning semantically meaningful weights to the proposed neighbors.

\subsection{Learned Attention Temperatures} 
The $\beta$ parameters in Eq. \ref{eq:exp_affinity}, which control the sharpness of the cross-attention function, are learned independently for each cross-attention layer in our network. To ensure training stability, we optimize the logarithm of $\beta$ rather than the raw values. In Table \ref{tab:beta_values}, we reported the learned $\beta$ values for AttentionViG-S after training on ImageNet-1K. 

We observed that the model favors sharper mixing in early and mid-level layers, requiring stronger alignment between node queries and neighbor keys for interaction. Interestingly, the very first cross-attention layer is an exception, exhibiting a relatively softer affinity. This may indicate a need for broader context aggregation at the input stage before more selective mixing is applied in subsequent layers. In contrast, later layers again shift toward softer affinities, suggesting a progressive relaxation in mixing strictness as the model deepens.     

\vspace{-1mm}
\subsection{Ablation Studies}
\vspace{-1mm}

\label{sec:ablation}

\begin{table}[!t]
    \centering
    \captionsetup{font=scriptsize}
    \scriptsize
    \renewcommand{\arraystretch}{0.80}
    \setlength{\tabcolsep}{5pt} 
    
    \begin{minipage}{0.29\textwidth}
        \centering
        \caption{Learned $\beta$ values for cross-attention layers across scales.}
        \label{tab:beta_values} 
        \begin{tabular}{@{}c cc@{}}
            \toprule
            \textbf{Stage} & \textbf{Attn. L1} & \textbf{Attn. L2} \\
            \midrule
            1 & 6.79 &  29.88 \\
            2 & 23.72 & 10.96 \\
            3 & 6.63 & 7.08 \\
            4 & 5.33 & 4.92 \\
            \bottomrule
        \end{tabular}
    \end{minipage}\hfill
    \begin{minipage}{0.31\textwidth}
        \centering
        \caption{Impact of aggregation functions on top-1 accuracy of vanilla ViG \cite{vig}.}
        \label{tab:ablation_aggregation}
        \begin{tabular}{@{}l c c@{}}
            \toprule
            \textbf{Aggregation} & \textbf{Acc. (\%)} & \textbf{GFLOPs} \\
            \midrule
            GIN \cite{gin} & 72.8 & 1.3 \\
            Max-Rel. \cite{mr} & 73.9 & 1.3 \\
            GraphSAGE \cite{graphsage} & 74.0 & 1.6  \\
            EdgeConv \cite{edgeconv} & 74.3 & 2.4 \\
            \textbf{Cross-Attn.} & \textbf{74.3} & \textbf{1.6} \\
            \bottomrule
        \end{tabular}
    \end{minipage}\hfill
    \begin{minipage}{0.31\textwidth}
        \centering
        \caption{Impact of the attention nonlinearity on top-1 accuracy of AttentionViG.}
        \label{tab:ablation_nonlinearity}
        \begin{tabular}{@{}l c@{}}
            \toprule
            \textbf{Nonlinearity} & \textbf{Acc. (\%)} \\
            \midrule
            Softmax & 80.8 \\
            Exp. affinity (1$/\#$neigh.) & 80.7 \\
            \textbf{Exp. affinity (no norm)} & \textbf{81.3} \\
            \bottomrule
        \end{tabular}
    \end{minipage}
    \vspace{-4mm}
\end{table}

We conducted ablation studies on ImageNet-1K \cite{imagenet} to evaluate classification performance. Integrated into Vanilla ViG \cite{vig} graph construction policy as the aggregation function, our cross-attention method outperforms Max-Relative, GIN, and GraphSAGE, and matches EdgeConv while using only 66\% of its FLOPs.

In Table \ref{tab:ablation_nonlinearity}, we evaluate the effect of different attention nonlinearities. The proposed exponential affinity yields a +0.5\% performance gain over softmax. Unlike softmax, which enforces competition among neighbors, the exponential function (see Eq. \ref{eq:exp_affinity}) assigns attention scores independently based on similarity, enabling more flexible and expressive aggregation. This result suggests a broader implication: enforcing competition among attended regions, as softmax does, may limit the expressivity of attention mechanisms in visual tasks. Additionally, normalizing attention scores by the number of neighbors leads to a 0.6\% performance drop, likely due to oversmoothing.

\vspace{-3mm}
\section{Conclusion} 
\label{sec:conclusion} 
We have proposed a cross-attention-based node-neighbor feature aggregation method for ViGs. Unlike prior work, our approach learns the optimal contribution of each neighbor independently of the graph construction policy, making it more robust to imperfections in such policies. Additionally, we have introduced AttentionViG, a hybrid CNN-GNN backbone that integrates the proposed aggregation function with inverted residual blocks to enhance efficiency while preserving expressive power. 

Our extensive experiments on classification, detection, and segmentation show that AttentionViG delivers competitive performance across scales compared to SOTA hybrid and token-mixing models. Its cross-attention aggregation proves effective for structured visual data, with solid results in both classification and dense prediction tasks. 

While our focus is image recognition, the proposed aggregation is broadly applicable to graph-based learning. Future work may extend it to video understanding, point cloud processing, and biological networks, where adaptive, structure-aware neighbor interactions are crucial. 

\section*{Acknowledgments}
This work was supported in part by the National Science Foundation (NSF) under grant ~CCF-2531882 and the Office of Naval Research (ONR) under grant~N00014-22-1-2254. 

\bibliographystyle{unsrtnat}
\bibliography{main}
\end{document}